\documentclass[runningheads]{llncs}
\usepackage{graphicx}
\usepackage{mathtools}
\usepackage{amsmath}
\usepackage{amssymb}
\usepackage{multirow}
\usepackage{booktabs}
\usepackage{graphicx}
\usepackage{overpic}
\usepackage{bbm}
\usepackage{xcolor}
\usepackage{mathrsfs}

\graphicspath{{figures/}}
\newcommand{\para}[1]{\vspace{1.0mm}\noindent\textbf{{#1}}~}

\begin{document}

\title{ToothInpaintor: Tooth Inpainting from Partial 3D Dental Model and 2D Panoramic Image}

\titlerunning{ToothInpaintor: Tooth Inpainting from Partial 3D Dental Model and 2D Panoramic Image}

\author{Yuezhi Yang\inst{1}
   , Zhiming Cui\inst{1,2}, Changjian Li\inst{3}, Wenping Wang\inst{1}}

\authorrunning{Y. Yang et al.}
%
\institute{ Department of Computer Science, The University of Hong Kong, Hong Kong \email{yzyang@cs.hku.hk} \and   Shanghai United Imaging Intelligence Co., Ltd., China \and
IPAB, School of Informatics, The University of Edinburgh, UK}

\maketitle              
\begin{abstract}

In orthodontic treatment, a full tooth model consisting of both the crown and root is indispensable in making the treatment plan. 
However, acquiring tooth root information to obtain the full tooth model from CBCT images is sometimes restricted due to the massive radiation of CBCT scanning. Thus, reconstructing the full tooth shape from the ready-to-use input, e.g., the partial intra-oral scan and the 2D panoramic image, is an applicable and valuable solution. 
In this paper, we propose a neural network, called \emph{ToothInpaintor}, that takes as input a partial 3D dental model and a 2D panoramic image and reconstructs the full tooth model with high-quality root(s).
Technically, we utilize the implicit representation for both the 3D and 2D inputs, and learn a latent space of the full tooth shapes. At test time, given an input, we successfully project it to the learned latent space via neural optimization to obtain the full tooth model conditioned on the input. To help find the robust projection, a novel adversarial learning module is exploited in our pipeline.
We extensively evaluate our method on a dataset collected from real-world clinics. The evaluation, comparison, and comprehensive ablation studies demonstrate that our approach produces accurate complete tooth models robustly and outperforms the state-of-the-art methods.

\end{abstract}

\section{Introduction}

In computer-aided orthodontic dentistry, a full tooth shape with roots is indispensable in evaluating the past treatment, measuring the teeth movement, and making the new treatment plan\cite{hu2009relationships,liang2020x2teeth}. Cone-bean computed tomography (CBCT) images are the only data modality containing 3D tooth root information.
However, considering the massive radiation of CBCT scanning, utilizing a CBCT to acquire the full tooth shape of the patient in orthodontic treatment is not allowed in many countries in the world. 
Thus, reconstructing complete 3D tooth shape from the more accessible data modalities, i.e., the 2D panoramic image and the partial intra-oral dental model, is an applicable and promising direction.

\begin{figure*}[!t]
    \centering
    \begin{overpic}[width=\linewidth]{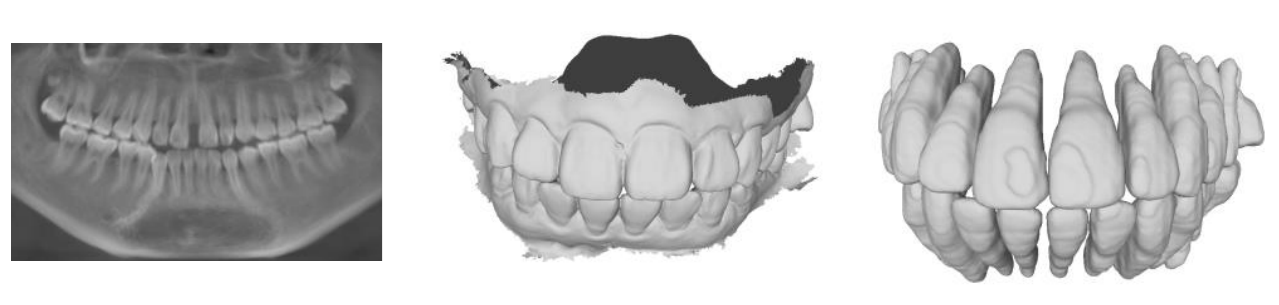}
        \put(13, -1) {\footnotesize (a)}
        \put(48, -1) {\footnotesize (b)}
        \put(82.3, -1) {\footnotesize (c)}
    \end{overpic}
    \vspace{-6mm}
    \caption{Given (a) a panoramic image and (b) several partial dental models as inputs, (c) the faithful full tooth models are expected to be predicted.}
    \label{fig:teaser}
    \vspace{-4mm}
\end{figure*}

It is, however, a challenging task due to the following facts. First, the panoramic image only contains the projected tooth information (Fig. \ref{fig:teaser}(a)), which has the inherent ambiguity from 2D to 3D. And second, the intra-oral scan contains the crown shape solely (Fig. \ref{fig:teaser}(b)). 
Previous works \cite{barone2015geometrical,pei2011personalized,barone20143d} have been explored to utilize template-fitting based methods to deform templates to match the tooth contour and the crown shape. However, these methods require a set of pre-defined tooth templates, which cannot represent the varying tooth shape appearance flexibly. Recently, many learning-based methods have been proposed for general 3D shape generation and completion \cite{park2019deepsdf,dai2017shape}. But these methods only consider the partial 3D information, which is not applicable for our specific task where the panoramic image provides the trusted tooth contour as an additional reference.

In this paper, we present a novel learning-based method for tooth reconstruction from a partial 3D dental model and a 2D panoramic image. Briefly, inspired by the successful attempts of the implicit representation \cite{park2019deepsdf} in deep learning, we represent both the 2D and 3D tooth information via the corresponding signed distance function (SDF) fields. With the SDF representation, we then build a latent space of full tooth models using an auto-decoder neural network, where various tooth types and shapes are encoded and each code corresponds well to a high-quality full tooth shape.At test time, given the 3D SDF of the partial dental model and the 2D SDF of the panoramic image as inputs, the goal is to project them into the latent space to find a faithful tooth shape code to derive the full model. This is achieved by neural optimization via the auto-decoder network, where the network parameters are fixed and only the resulting code is optimized. Furthermore, to help maintain the global tooth shape during the code optimization, adversarial learning is introduced to assist the robust projection. The extensive experiments show that our method can reconstruct high-quality complete tooth models with the panoramic image and a partial dental model, which has significantly outperformed the state-of-the-art performance and offer the potential usability of our framework in real-world clinics.

\section{Method}
\label{sec:method}
\begin{figure*}[!t]
    \centering
    \begin{overpic}[width=\linewidth]{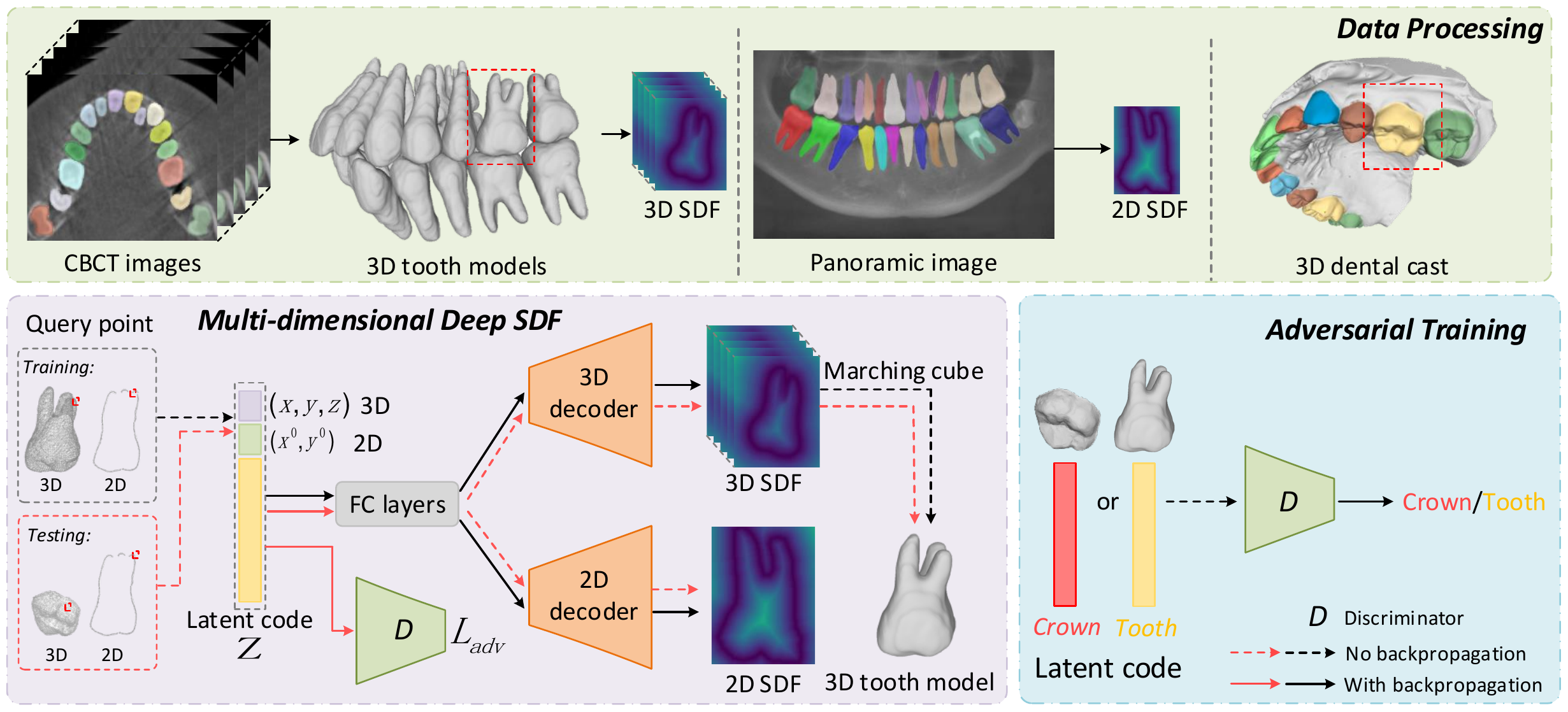}
    \end{overpic}
    \vspace{-7mm}
    \caption{The pipeline of \emph{ToothInpaintor}, which consists of the data processing (Sec. \ref{subsec:data_preprocessing}), mutli-dimensional deep SDF and the adversarial learning (Sec. \ref{subsec:deep_sdf}) modules. Note here, a dotted line means no back-propagation in the corresponding data flow, as illustrated in the legend.}
    \label{fig:pipeline}
    \vspace{-4mm}
\end{figure*}

Fig. \ref{fig:pipeline} shows an overview of our \emph{ToothInpaintor}, which consists of the data processing, multi-dimensional SDF, and adversarial learning modules. We elaborate on the pipeline in this section. 

\subsection{Data Processing}
\label{subsec:data_preprocessing}
As shown in Fig. \ref{fig:pipeline}, given the inputs of CBCT images, the panoramic image, and the intra-oral scan, we process them to obtain the required data as follows. Firstly, we adopt ToothNet \cite{cui2019toothnet} to accurately segment tooth individuals from CBCT images and reconstruct them as the ground truth full tooth shapes. We then utilize TSegNet \cite{cui2020tsegnet} to faithfully segment tooth crown from the dental model. 
Finally, Mask R-CNN \cite{he2017mask} is employed to segment the individual tooth from the panoramic image.
To map the panoramic image back to the tooth models, we first calculate the biquadratic curve of the tooth center points to fit a dental arch curve. Then, all 2D tooth centers from the panoramic image are obtained and fitted to the dental arch curve so that each tooth contour can align well to its corresponding tooth model. To better calculate the SDF field, we normalize the extracted full tooth individuals within a unit ball, and roughly scale and place the partial crown model to fit a full tooth shape.

\subsection{Multi-dimensional Deep SDF}
\label{subsec:deep_sdf}

In our approach, we utilize the signed distance function (SDF) to represent both the 2D and 3D tooth shapes. Take the 3D tooth shape as an example, given a spatial point, its SDF value indicates the distance to the closest surface and the sign refers to whether the point is inside (negative) or outside (positive) of the watertight surface. With the SDF representation, the tooth model can be directly instantiated by Marching Cubes \cite{lorensen1987marching}. 

Since we have both the 3D tooth and 2D contour shapes, we exploit multi-dimensional SDFs as in the following.

\para{3D SDF.}
Inspired by DeepSDF \cite{park2019deepsdf}, given a tooth shape, we prepare a set of pairs $X$ consists of the 3D point samples and their SDF values:
\begin{equation}
X = \{((x,y,z), s): s = SDF((x,y,z))\}.
\end{equation}

Intuitively, having these data pairs, we can train a neural network to produce the SDF value given any spatial query point for a specific tooth model. However, we want a neural network that can represent various tooth shapes, and embed them together in a low-dimensional latent space. Thus, as shown in Fig. \ref{fig:pipeline}, we utilize a shape code $z$ (256-$dim$ vector) as the additional input to the network and map it to the desired shape represented by the continuous SDF field using the network $f_\theta((x,y,z), z_{i})$, defined as:
\begin{equation}
    f_{\theta}((x,y,z), z_i) \approx  SDF^i((x,y,z)) 
\end{equation}
where $x$ is an arbitrary spatial point, $z_i$ refers to the shape code of the $i$th shape that needs to be learned by the network, and $SDF^i((x,y,z))$ is the ground truth SDF value of the $i$th shape. Note here, $f_\theta$ is an auto-decoder network  and composed of several Fully Connected (FC) layers to produce the SDF values.

\para{2D SDF.}
Recall that, the 2D tooth contour shape serves as an important shape reference in our task, and we further use the 2D SDF derived from the tooth contour.
Similarly, we calculate a 2D SDF field from each tooth contour, where the value of each pixel indicates the signed distance to the iso zero-contour. 
To incorporate the 2D SDF input, as can be seen from Fig. \ref{fig:pipeline}, we simply concatenate the 2D \emph{pixel-level} query point $(x^0, y^0)$, the latent code $z$, and the 3D spatial point $(x,y,z)$ together. And our auto-decoder network has an extra output branch to predict the 2D SDF. In this way, the network can automatically build the intrinsic relationship between the 3D tooth shape and the projected 2D contour.

\para{Loss Function.}
Under the configuration of deep SDF learning, the network parameters and latent code $z$ should be optimized jointly, and we supervise the prediction of the 2D and 3D SDFs. Additionally, we introduce a regularization term on the latent code to improve the generalization ability. All together, the loss function $\mathcal{L}$ is formulated as:
\begin{equation}
    \underset{\theta , z}{min} \mathcal{L} = \mathcal{L}^{3D}_{SDF} + \mathcal{L}^{2D}_{SDF} + \lambda  \left \| z \right \|_{2},
\end{equation}
where $\mathcal{L}^{3D}_{SDF}$ and $\mathcal{L}^{2D}_{SDF}$ refer to the L1 loss of the 3D and 2D SDFs prediction, respectively. 
And $\lambda$ is a balancing weight that is set as $0.0001$ in all experiments. 

\subsection{Shape Code Optimization with Adversarial Learning}

At the testing stage, the goal is to project the inputs (i.e., the crown model and the panoramic image) to the learned latent space to find a correct code that encodes the full tooth shape respecting the inputs. We achieve it using neural optimization. Specifically, we fix the network parameters since it has captured the tooth shape priors when building the latent space, and only update the code given the supervision of the crown shape and the contour shape. See Fig. \ref{fig:pipeline} for an illustration, where the red lines present the data flow at the testing time and dotted lines mean no gradient back-propagation. Yet, when solely supervising the prediction of the SDFs to optimize the code, the crown shape tends to dominate the optimization so that the resulting code cannot maintain the global tooth shape well, e.g., with irregular tooth roots. To resolve this issue, we further propose adversarial learning in this optimization step.

\para{Adversarial Learning.}
To use the discriminator, we should first train it at the training time. Specifically, we have the code $z_{full}$ representing the full tooth shape, and also optimize the partial shape code $z_{partial}$ by feeding the network with the crown model. To learn $z_{partial}$, we disable the gradient back-propagation of the auto-decoder network parameters when supervising the reconstruction of the partial shape to update the code solely. As illustrated in Fig. \ref{fig:pipeline}, we take $z_{full}$ as the positive example, and  $z_{partial}$ as the negative example to train the discriminator.
Note here, the discriminator loss is only used to optimize itself instead of the learning of the code and auto-decoder network parameters.
After training, the discriminator has the ability to tell whether a shape code represents a regular full shape or not.

\para{Testing Stage Optimization.}
With the learned network $f_\theta$ and the discriminator, we optimize the resulting shape code $\widehat{z}$ by supervising the reconstruction and the discriminator losses, defined as:
\begin{equation}
    \widehat{z} = \underset{z}{argmin} (\mathcal{L}^{3D}_{SDF} + \mathcal{L}^{2D}_{SDF} + \lambda  \left \| z \right \|_{2} + \mathcal{L}_{adv}),
\end{equation}
where the $\mathcal{L}^{3D}_{SDF}$ and the $\mathcal{L}^{2D}_{SDF}$ are the same L1 loss as in the training stage, but the only difference is that the 3D SDF comes from the partial crown model. And $\mathcal{L}_{adv}$ is the common adversarial loss.

\para{Stopping Criteria.} 
The well-learned discriminator is the key module to help maintain the global tooth shape in neural optimization. Hence, instead of stopping the optimization by checking the SDF reconstruction error, e.g., lower than a proper threshold, we refer to the discriminator loss. That is, given the supervision of positive labels, in case the discriminator converges to a low error, we then stop the optimization. In this way, the SDF loss reaches an acceptable low value indicating the satisfactory reconstruction of the inputs, while the resulting code successfully fools the discriminator demonstrating that the code represents a regular full tooth shape.

Once we have the optimized latent code $\widehat{z}$, we directly feed forward the network to calculate the 3D SDF value of each query point in a spatial grid with $512^3$ resolution. Then, the predicted complete tooth shape can be reconstructed by Marching Cubes \cite{lorensen1987marching}.

\para{Post-nonrigid Deformation.} 
The network cannot always guarantee to reproduce the input crown shape due to the optimization nature, we thus exploit a post-nonrigid deformation step to further refine the tooth crown shape to have the geometric details from the crown input. Specifically, laplacian surface editing \cite{sorkine2004laplacian} is employed with a set of deformation handles built by nearest neighbor searching between the predicted tooth model and the crown model. Finally, we take the tooth models after non-rigid deformation as our final results.  

\subsection{Implementation Details and Network Training}
Our framework was implemented using PyTorch, and we used Adam optimizer to train the framework, where the learning rates are set as $10^{-3}$ and $10^{-4}$, respectively. In total, we trained the whole network 1000 epochs in the training stage for about 12 hours using a Nvidia 1080Ti GPU.
The auto-decoder network has 7 FC layers as backbone and 7 and 5 FC layers for the following 3D and 2D branches, respectively. The discriminator has 5 FC layers.

\section{Experiments}

\para{Dataset.}
To train and test our approach, we have collected a dataset from real-world clinics, where different data modalities can be obtained by scanning.

In total, there are 135 paired CBCT images, panoramic images, and intra-oral scans in our dataset, where the resolution of the CBCT images and panoramic images are $0.35mm$ and $0.1mm$, respectively.
We randomly split it into 100, 10, and 25 for training, validating, and testing, respectively.

\para{Evaluation Metrics.}
To quantitatively evaluate the performance of our method, we make use of several metrics to measure the reconstruction accuracy. Concretely, Chamfer distance (CD), Hausdorff distance (HD), as well as average surface distance (ASD) are chosen and are reported in Table \ref{tab:ablation}.

\subsection{Ablation Study}
\label{subsec:ablation}

\begin{table}[!t]
\caption{Quantitative results of alternative ablation networks. The smaller the value, the better the reconstruction accuracy.}
\vspace{-2mm}
\centering
\setlength{\tabcolsep}{4mm}\begin{tabular}{l|cccccc}
\hline
\toprule[1pt]
 Method & CD$\downarrow$ & HD$\downarrow$ & ASD$\downarrow$ \\ \hline
 bNet   & 0.96 & 2.56 & 0.47 \\
 bNet-P & 0.79 & 2.39 & 0.37 \\ 
 bNet-P-D   & 0.78 & 2.38 & 0.36  \\\hline
 FullNet       & \textbf{0.75} & \textbf{2.36} & \textbf{0.33}\\ \bottomrule[1pt] \hline
\end{tabular}
\label{tab:ablation}
\vspace{-4mm}
\end{table}

To demonstrate the efficacy of the key network components or loss terms, we have conducted additional experiments with alternative configurations. Specifically, we first build the baseline network denoted as bNet, with only the 3D full tooth shapes to build the manifold and the partial scan solely to optimize a code to produce the full tooth shape. All alternative configurations are derived by augmenting bNet with one network component or loss term and trained using the same training dataset from scratch.

\begin{figure*}[!t]
    \centering
    \begin{overpic}[width=\linewidth]{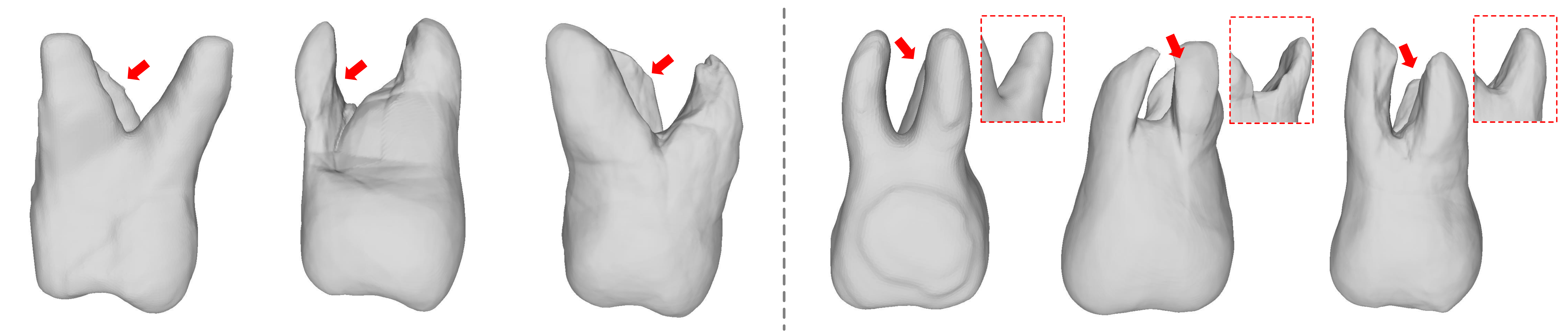}
        \put(4, -1.5) {\scriptsize (a) GT}
        \put(20.0, -1.5) {\scriptsize (b) bNet}
        \put(36, -1.5) {\scriptsize (c) bNet-P}
        \put(54, -1.5) {\scriptsize (d) GT}
        \put(69, -1.5) {\scriptsize (e) bNet-P}
        \put(86, -1.5) {\scriptsize (f) bNet-P-D}
    \end{overpic}
    \vspace{-5mm}
    \caption{The visual results of ablation experiments. The left half shows the results w/wo 2D SDF input, and the right half shows the results w/wo adversarial learning.}
    \label{fig:ab_2dSDF_D}
    \vspace{-4mm}
\end{figure*}
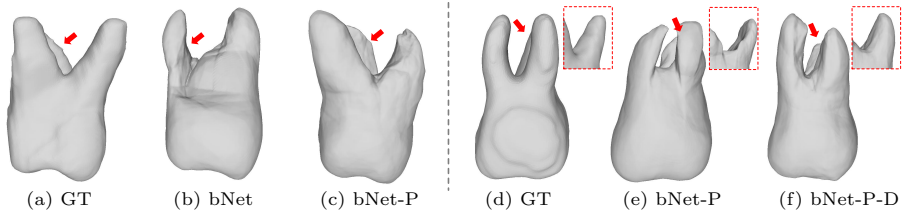

\para{Benefits of 2D SDF from Panoramic Image.}
Reconstructing the full 3D tooth shape from the partial crown model is a highly ill-posed problem, but the 2D contour from the panoramic image can alleviate the ambiguity to some extent. To validate its effectiveness, we augment bNet with the extra 2D SDF input, denoted as bNet-P. The quantitative results are presented in Table \ref{tab:ablation}, where compared to bNet, the CD error drops from 0.96 to 0.79 (i.e., 0.17 increasing). The improvement is consistent with the qualitative comparison in the left half of Fig. \ref{fig:ab_2dSDF_D}. It can be seen that guided by the tooth contour from the 2D SDF, bNet-P produces a tooth with correct root numbers and a reasonable shape.

\para{Benefits of Adversarial Learning.} Adversarial learning plays an important role in our framework, to validate, we further augment bNet-P with adversarial learning (denoted as bNet-P-D) to validate its efficacy. Statistically, although bNet-P-D gains only a little improvement (e.g., 0.01) in terms of all three metrics as can be seen from Tab. \ref{tab:ablation}, the visual results in the right half of Fig. \ref{fig:ab_2dSDF_D} can reveal the benefit more clearly. Since the panoramic image only provides the projected contour shape, for molar teeth with three roots, only a tiny part of the third root can be seen due to occlusion. Thus, without adversarial learning to maintain the global tooth shape, bNet-P generates the unsatisfactory result (Fig. \ref{fig:ab_2dSDF_D}(e)) with irregular roots, especially the sunken third root. However, the root with a relatively small area contributes a tiny portion of the error. Instead, bNet-P-D produces almost perfect result (Fig. \ref{fig:ab_2dSDF_D}(f)).

\begin{figure*}[!t]
    \centering
    \begin{overpic}[width=0.75\linewidth]{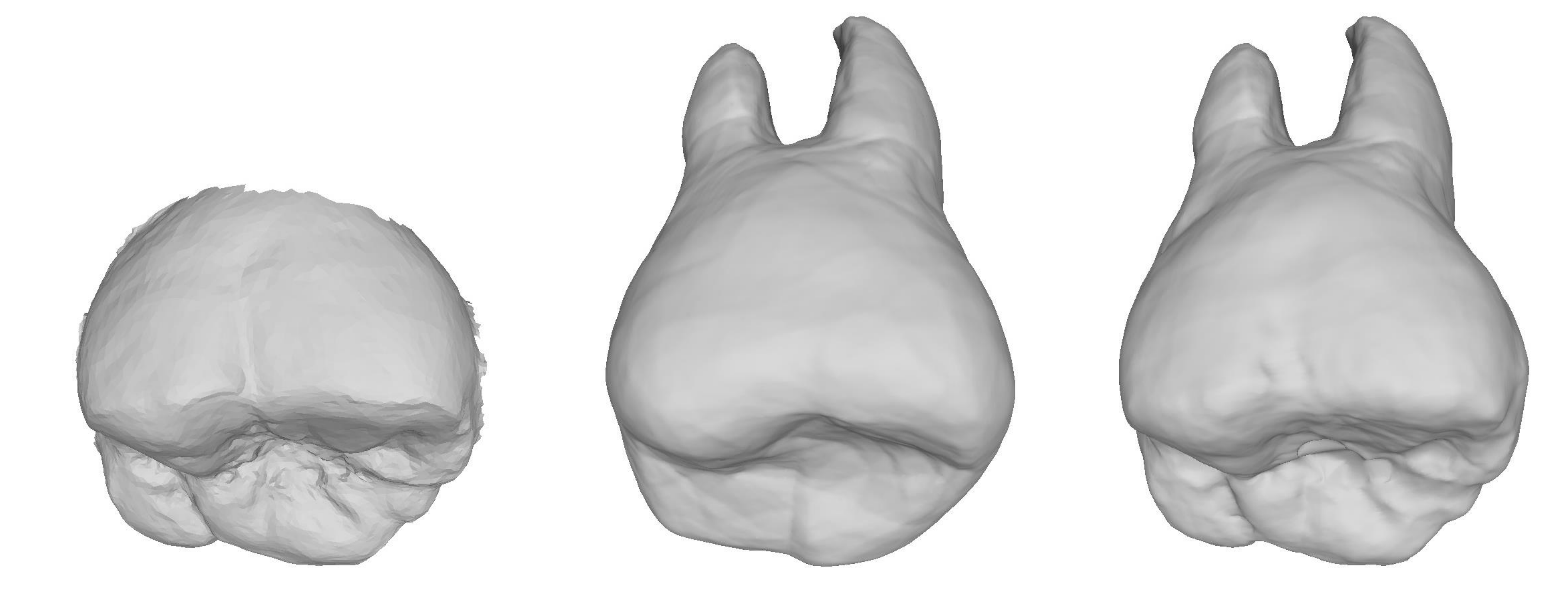}
        \put(7, -2) {\scriptsize (a) dental crown}
        \put(43, -2) {\scriptsize (b) bNet-P-D}
        \put(78, -2) {\scriptsize (c) FullNet}
    \end{overpic}
    \vspace{-1mm}
    \caption{The visual comparison of tooth inpainting with (c) or without (b) the post-deformation, compared to (a) the input dental crown.} 
    \label{fig:ab_deform}
    \vspace{-4mm}
\end{figure*}

\begin{figure*}[!t]
    \centering
    \begin{overpic}[width=\linewidth]{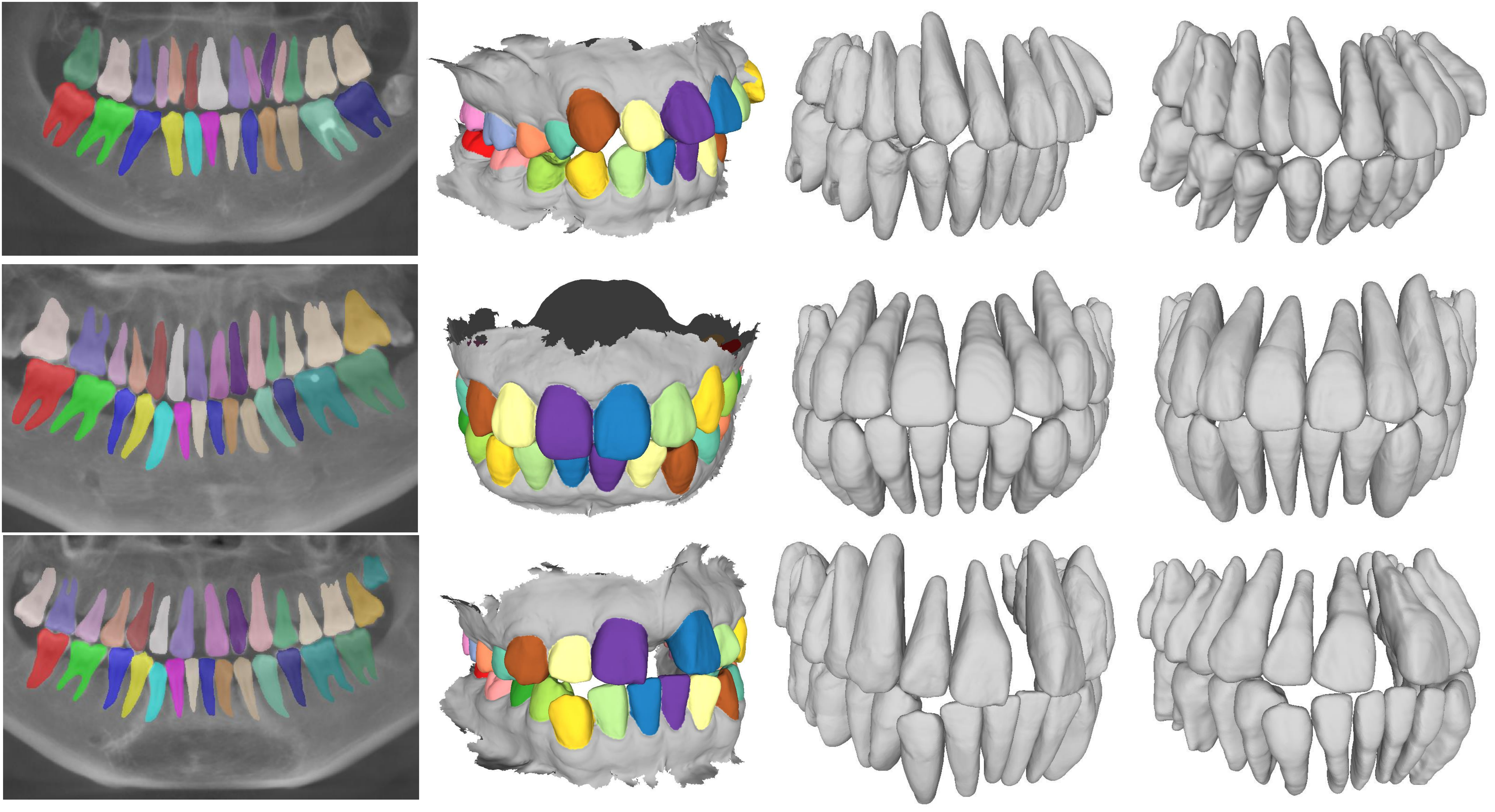}
        \put(3, -3) {\footnotesize Panoramic image}
        \put(29, -3) {\footnotesize Partial dental model}
        \put(59, -3) {\footnotesize Our result}
        \put(81, -3) {\footnotesize Ground truth}
    \end{overpic}
    \vspace{-2mm}
    \caption{Three typical examples of the tooth inpainting from the 2D panoramic image and 3D partial dental model.}
    \label{fig:result_gallery}
    \vspace{-3mm}
\end{figure*}

\para{Benefits of Post-deformation.} To validate the effectiveness of the post-deformation step, we augment bNet-P-D with this step as our FullNet. As can be seen from Tab. \ref{tab:ablation}, all the metrics are consistently improved, which is consistent with the visual results in Fig. \ref{fig:ab_deform}, where the geometric details of the crown part are kept. The statistical and visual results demonstrate the efficacy of our FullNet, which offers the potential usability of our framework in real-world clinical scenarios. More results can be seen in the result gallery of Fig. \ref{fig:result_gallery}.

\subsection{Comparison with State-of-the-art Methods}
\label{subsec:comparison}

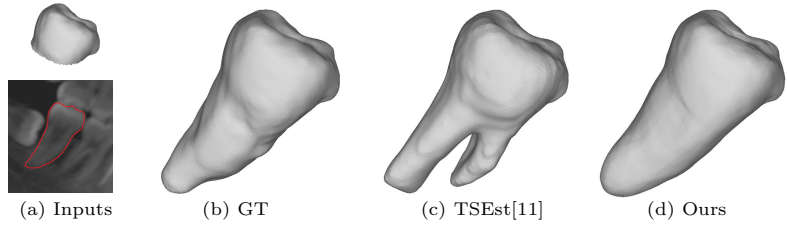
\begin{figure*}[!t]
    \centering
    \begin{overpic}[width=0.85\linewidth]{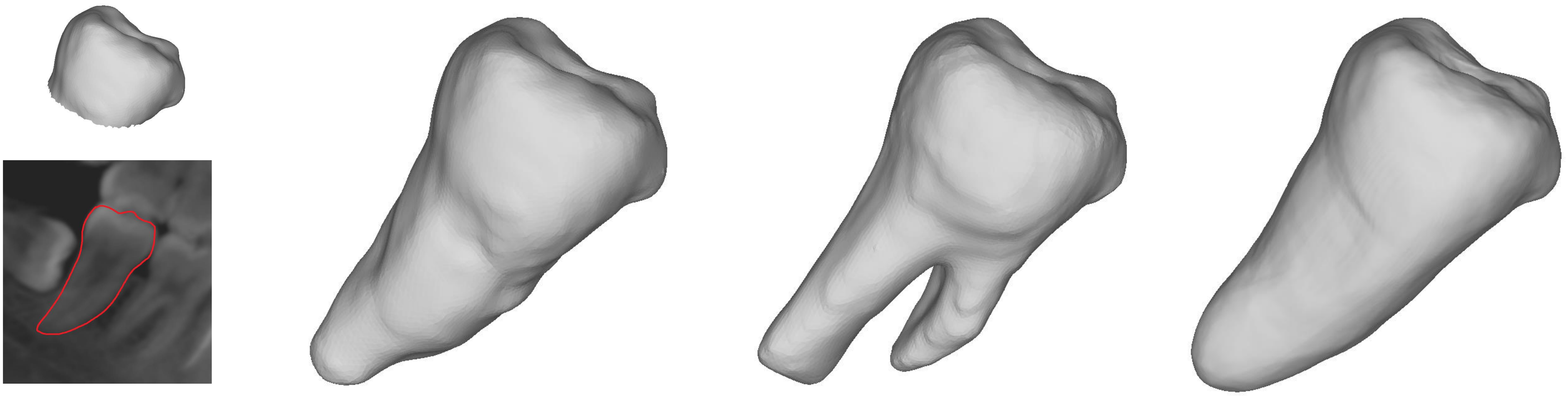}
        \put(1.5, -2) {\scriptsize (a) Inputs}
        \put(25, -2) {\scriptsize (b) GT}
        \put(53, -2) {\scriptsize (c) TSEst\cite{pei2011personalized}}
        \put(82, -2) {\scriptsize (d) Ours}
    \end{overpic}
    \vspace{-1mm}
    \caption{Visual comparison with TSEst  \cite{pei2011personalized}. }

    \label{fig:comparison}
    \vspace{-6mm}
\end{figure*}

Most of the state-of-the-art methods \cite{barone2015geometrical,pei2011personalized,barone20143d} adopt a template-fitting framework to reconstruct the complete tooth shape from the crown and panoramic image. To conduct a fair comparison, we implement \cite{pei2011personalized} (denoted as TSEst) to present the visual results. Notably, although the overall tooth shape looks plausible, the inherent drawback of template fitting is that it cannot automatically deviate the template to match the input faithfully. As shown in Fig. \ref{fig:comparison}, it happens to have a molar tooth with only one root as represented in the panoramic image. Not surprisingly, TSEst produces a tooth with two roots due to the pre-selected tooth template even if it disagrees with the contour shape. Instead, we automatically obtain a superior result similar to the ground truth.

\section{Conclusion}
In this paper, we proposed a neural solution to reconstruct a full tooth shape from a partial dental model and a panoramic image.
Our method is fully automatic without any tooth template and user interaction. It produces promising results by first building a faithful complete tooth shape latent space and then projecting the inputs to find a full tooth shape code that respects the inputs. 
We have evaluated our approach both qualitatively and quantitatively, and compared it against state-of-the-art methods, where our approach produces superior results and outperforms others. The outstanding results offer the potential usability of our framework in real-world clinical scenarios.

\bibliographystyle{splncs04}
\bibliography{toothRecons}

\end{document}